\documentclass[journal]{IEEEtran}

\usepackage{bbding}
\usepackage{makecell}
\usepackage{amsmath,amssymb}
\usepackage{graphicx}
\usepackage{setspace}
\usepackage{longtable}
\usepackage{colortbl,booktabs}
\usepackage{tabularx,multirow}
\usepackage{xcolor}
\usepackage{tikz}
\usepackage{multirow}
\usepackage{comment}
\graphicspath{{./pic/}}
\usepackage{color}
\usepackage{amsmath, amssymb, amsfonts}
\usepackage[linesnumbered, ruled, vlined]{algorithm2e}
 
\newlength\myindent
\usepackage{mathrsfs} 

\usepackage{array}  
\usepackage{booktabs}

\definecolor{lightgray}{gray}{0.9}
\definecolor{lightblue}{rgb}{0.68, 0.85, 0.9}

\definecolor{lightpurple}{rgb}{0.85, 0.75, 0.95}
\definecolor{lightorange}{rgb}{1.0, 0.9, 0.8}

\usepackage{hyperref}
\usetikzlibrary{fadings}
\usetikzlibrary{patterns}
\usetikzlibrary{shadows.blur}
\usetikzlibrary{arrows,shapes,chains}
\usepackage[numbers]{natbib}
\makeatletter
\def\hlinewd#1{%
	\noalign{\ifnum0=`}\fi\hrule \@height #1 \futurelet
	\reserved@a\@xhline}
\makeatother

\begin{document}

\title{PointSAM: Pointly-Supervised Segment Anything Model for Remote Sensing Images}

\IEEEtitleabstractindextext{
\begin{abstract}
Segment Anything Model (SAM) is an advanced foundational model for image segmentation, which is gradually being applied to remote sensing images (RSIs). Due to the domain gap between RSIs and natural images, traditional methods typically use SAM as a source pre-trained model and fine-tune it with fully supervised masks. Unlike these methods, our work focuses on fine-tuning SAM using more convenient and challenging point annotations. Leveraging SAM’s zero-shot capability, we adopt a self-training framework that iteratively generates pseudo-labels. However, noisy labels in pseudo-labels can cause error accumulation. To address this, we introduce Prototype-based Regularization, where target prototypes are extracted from the dataset and matched to predicted prototypes using the Hungarian algorithm to guide learning in the correct direction. Additionally, RSIs have complex backgrounds and densely packed objects, making it possible for point prompts to mistakenly group multiple objects as one. To resolve this, we propose a Negative Prompt Calibration method, based on the non-overlapping nature of instance masks, where overlapping masks are used as negative signals to refine segmentation. Combining these techniques, we present a novel pointly-supervised segment anything model, PointSAM. We conduct experiments on three RSI datasets, including WHU, HRSID, and NWPU VHR-10, showing that our method significantly outperforms direct testing with SAM, SAM2, and other comparison methods. Additionally, PointSAM can act as a point-to-box converter for oriented object detection, achieving promising results and indicating its potential for other point-supervised tasks. The code is available at \url{https://github.com/Lans1ng/PointSAM}.

\end{abstract}

}

\markboth{Journal of \LaTeX\ Class Files,~Vol.~14, No.~8, August~2015}%
{Shell \MakeLowercase{\textit{et al.}}: Bare Demo of IEEEtran.cls for IEEE Journals}
	\author{Nanqing~Liu, Xun~Xu, Yongyi~Su, Haojie~Zhang, Heng-Chao Li
	\IEEEcompsocitemizethanks{
		\IEEEcompsocthanksitem 

  Nanqing Liu (lansing163@163.com) is with School of Information Science and Technology, Southwest Jiaotong University, Chengdu, China, and also with I2R, A*STAR, Singapore 138632. Xun Xu (xux@i2r.a-star.edu.sg) is with I2R, A*STAR, Singapore 138632.
  Yongyi Su (eesuyongyi@mail.scut.edu.cn) is with South China University of Technology, and also with I2R, A*STAR, Singapore 138632. Haojie Zhang is with South China University of Technology.
  Heng-Chao Li (lihengchao\_78@163.com) is with School of Information Science and Technology, Southwest Jiaotong University, Chengdu, China.

}
	\thanks{Manuscript revised \today.}}
\maketitle
\IEEEdisplaynontitleabstractindextext


\section{Introduction}
Foundation models are versatile, large-scale models designed for a wide range of tasks and applications. They have demonstrated exceptional performance in areas such as natural language processing (e.g., BERT\cite{bert} and GPT-3\cite{gpt3}) and multimodal tasks (e.g., CLIP\cite{clip} and ALIGN \cite{align}). Recently, Segment Anything Model (SAM) \cite{sam,sam2} was introduced as a foundation model specifically for image segmentation. Trained on a billion-scale dataset of masks and prompts, SAM can be applied to various downstream tasks requiring promptable segmentation, including healthcare \cite{ma2024segment,huang2024segment}, autonomous driving \cite{shan2023robustness}, and remote sensing \cite{chen2024rsprompter,wang2024samrs,10412208}.

Despite SAM’s strong zero-shot capabilities, challenges persist in handling out-of-distribution (OOD) data and domain shifts in remote sensing images (RSIs). Many categories in RSIs are not represented in SAM’s training data. Furthermore, RSIs are typically captured from aerial or satellite perspectives and differ significantly from natural images. Consequently, recent studies \cite{chen2024rsprompter,ding2024adapting,10315957,pu2024classwise} have focused on how to fine-tune SAM for specific tasks. For example, RS-Prompter \cite{chen2024rsprompter} uses queries or anchors as prompts to guide SAM’s mask decoder for instance segmentation. Similarly, SAM-CD \cite{ding2024adapting} employs FastSAM’s encoder and introduces adapters for fine-tuning in change detection tasks.

\begin{figure}[tp]
	\centering
\resizebox{0.95\linewidth}{!}{\input{pic/intro.tex}}
	\caption{(a) Training pipeline of vanilla SAM. (b) Training pipeline of self-training based pointly-supervised SAM. \textit{Sup.} means supervise.
	}
        \vspace{-0.3cm}
	\label{fig:intro}
\end{figure}

While these methods achieve promising results, they rely on full mask annotations, which are difficult and time-consuming to obtain. To address this, recent approaches \cite{wesam,cat-sam,tang2024bootstrap} have focused on label-efficient strategies for SAM. WeSAM\cite{wesam} and SlotSAM\cite{tang2024bootstrap} use self-training \cite{st} with weak labels, such as points and boxes, to generate pseudo-labels, enabling the network to predict complete masks. Cat-SAM\cite{cat-sam} adopts a few-shot learning approach, fine-tuning SAM with box prompts for mask prediction. While box and coarse mask prompts have shown strong results, point-based supervision remains less effective. Moreover, point annotations are much cheaper than masks and boxes \footnote{https://cloud.google.com/ai-platform/data-labeling/pricing}, particularly for RSIs with numerous dense objects. Therefore, this paper aims to explore how to adapt SAM to RSIs using the most challenging yet cost-effective \textit{point} annotations. First, let us review how full mask annotations are used to fine-tune SAM for RSIs. As shown in Fig. \ref{fig:intro} (a), SAM takes point or box prompts as input to generate the predicted mask $\mathcal{M}_{pred}$, which is supervised by the ground truth (GT) mask $\mathcal{M}_{gt}$.
In contrast, the self-training-based method\cite{wesam,tang2024bootstrap} (depicted in Fig. 
\ref{fig:intro} (b)) only requires pseudo-labels generated by the model itself. Specifically, the input undergoes both weak and strong augmentations separately and is fed into the SAM's image encoder, resulting in $\mathcal{M}_{pred}$ and $\mathcal{M}^{\prime}_{pred}$, respectively. $\mathcal{M}_{pred}$ serves as a pseudo-label to constrain $\mathcal{M}^{\prime}_{pred}$, enabling iterative training. This method is feasible primarily due to the principles of source-free domain adaptation (SFDA)\cite{irg,liu2024clip,shot}. The core idea of SFDA is to improve model performance using unlabeled data from the target domain without requiring access to source domain data.


However, self-training often depends on the quality of pseudo-labels. If there is noise in the pseudo-labels, the model may overfit incorrect patterns. To address this, two common approaches are feature alignment \cite{stfar,actmad,yoo2024and} and logit regularization \cite{tribe,wesam}. However, the former requires access to the distribution of source data, which is impractical for SAM. The latter can also affect results if the prediction of anchor logits is inaccurate. In contrast to these methods, our approach aligns the features of the source and target models at the image encoder. Rather than performing simple image-level feature alignment, we map the corresponding prompt locations to the encoder features for instance-level feature alignment. Since object point labels are already annotated, we do not rely on inaccurate predicted logits for constraints or use source data information. Specifically, before beginning self-training, we first extract features for each instance from the target data using the source model. We then cluster these instances using the parameter-free clustering algorithm FINCH \cite{finch} and compute \textit{target prototypes} for all clusters. During self-training, we maintain a First-in-First-Out (FIFO) memory bank, which stores instance-level predicted features and similarly computes the \textit{predicted prototypes}. Since discrepancies between the number of targets and predicted prototypes may exist, direct correspondence cannot be established. To resolve this, we employ the Hungarian algorithm, which automatically matches these two types of prototypes and aligns them using a matching loss. We call this method \textbf{Prototype-based Regularization} (PBR).

\begin{figure}[tp]
	\centering
\resizebox{1.0\linewidth}{!}{\input{pic/intro_pos_neg.tex}}
        \vspace{-0.3cm}
	\caption{{Segmentation results on the NWPU VHR-10, WHU, and HRSID datasets. (a) Segmentation results using only positive prompts. (b) Segmentation results using both positive and negative prompts.}
	}

	\label{fig:intro_pos_neg}
\end{figure}

Moreover, RSIs are captured from overhead perspectives and contain densely detected objects and large-scale backgrounds, making points as prompts more semantically ambiguous because points lack boundary information. We tested RSIs on SAM’s demo website \footnote{https://segment-anything.com/demo}; as shown in Fig. \ref{fig:intro_pos_neg}  (a), the densely distributed instances in the image can cause the mask decoder to mistakenly interpret them as a single instance. However, after adding negative samples (shown in Fig. \ref{fig:intro_pos_neg} (b)), the remaining parts were effectively removed. Inspired by this, selecting appropriate locations for negative prompts is crucial. We thus propose a method for adaptively extracting negative prompts during training, called \textbf{Negative Prompt Calibration} (NPC). This procedure is based on a prior assumption: \textit{there is no overlap between predicted masks of different instances}. We first calculate the IoU between each instance and use other samples with an IoU above a certain threshold with respect to a given sample as candidate negative prompts. Then, we randomly select $k$ positive prompts to serve as negative prompts for the target sample. Finally, we input the new prompts into the mask decoder to obtain refined masks.

We integrate the above two methods into the self-training-based point-supervised framework, named PointSAM. We conduct experiments on three representative RSI datasets: NWPU VHR-10, WHU, and HRSID. The results demonstrate that our approach effectively adapts vanilla SAM to various RSI scenarios under point supervision. Additionally, we apply PointSAM as a bounding box generator in point-supervised oriented object detection tasks, indicating that this method can extend to other point-supervised applications. Our contributions are summarized as follows:

\begin{itemize} 
\item We propose Prototype-based Regularization (PBR), which extracts instance-level features from both the source and target models. By using non-parametric clustering, dynamically updating prototypes, and Hungarian matching, PBR prevents the model from learning in incorrect directions and improves model generalization.

\item We introduce Negative Prompt Calibration (NPC), which adaptively adjusts negative prompts during training. By using positive prompts from overlapping samples as negative prompts, NPC enhances the original SAM's predicted masks and achieves more accurate results in dense scenarios.

\item We are the first to leverage point annotations to fine-tune SAM for remote sensing images (RSIs). Extensive experiments on three datasets (NWPU VHR-10, WHU, and HRSID) validate the effectiveness of PointSAM, achieving significant improvements in point-supervised segmentation performance. Additionally, we extend PointSAM's application to bounding box generation for point-supervised oriented object detection, demonstrating its versatility and potential in point-based supervised learning tasks.
\end{itemize}







\section{Related Work}
\subsection{Segment Anything Model}
Segment Anything Model (SAM) \cite{sam,sam2} was developed by Meta AI, leveraging a large and diverse training dataset and a powerful neural network architecture to perform segmentation tasks on any image. By inputting points or bounding boxes as prompts, the desired instance masks can be obtained. To make it more suitable for various platforms or scenarios, some methods have been improved primarily in terms of speed and accuracy. To reduce the model complexity of SAM, researchers have focused on knowledge distillation and self-supervised techniques. For example, MobileSAM \cite{mobilesam} distills knowledge from the large image encoder ViT-H in the original SAM into a lightweight encoder. EfficientSAM \cite{efficientsam} employs a reconstruction self-supervised method using MAE to transfer knowledge to a smaller image encoder that replaces the original SAM encoder. To further enhance the segmentation accuracy, HQ-SAM \cite{sam_hq} introduces learnable High-Quality Output Tokens and their associated three-layer MLPs to correct the mask errors of SAM’s output tokens. Additionally, because SAM is category-agnostic, some methods \cite{ovsam,grounded_sam} have incorporated text models\cite{clip} to provide the masks with category information.

Thanks to SAM's strong zero-shot and generalization capabilities, it has also been successfully adapted to RSIs \cite{wang2024samrs,osco2023segment,ding2024adapting,yan2023ringmo,moghimi2024comparative}. Due to the semantic gap between RSIs and natural images, mainstream methods typically use SAM's encoder as a backbone and apply existing fine-tuning techniques, such as LoRA \cite{lora} and adapter methods. For example, TTP \cite{ttp} uses SAM's encoder as the backbone for change detection and fine-tunes with LoRA \cite{lora}. RSPrompter \cite{chen2024rsprompter} freezes some modules of SAM and uses adapters for instance segmentation. However, these methods require fully annotated data for fine-tuning. In contrast, our work focuses on fine-tuning SAM with minimal annotation costs, and we are the first to explore fine-tuning SAM using point annotations for RSIs.

\subsection{Point-based supervision}
Point annotations are often used to save on mask or box annotations. 
Compared to image-level annotations 
\cite{10580924}, it can indicate the object's location, providing stronger priors for subsequent processing and offering better practicality.
Point-supervised methods are widely applied in detection \cite{point2rbox,P2BNet,pointobb, pointobb2, pmho,cao2023p2rbox} or segmentation\cite{fan2022pointly,li2023point2mask,bearman2016s,cheng2022pointly,yuan2024beyond,10703135} tasks. For example, P2BNet \cite{P2BNet} uses Multiple Instance Learning (MIL) to select the box with the highest confidence from multiple boxes containing points. Point2Mask \cite{li2023point2mask} formulates the pseudo-mask generation from points as an Optimal Transport (OT) problem.
Unlike natural images, instances in RSIs are mostly smaller and more densely packed, making point annotations much more convenient for label generation. PointOBB \cite{pointobb} learns object scale and angle information through self-supervised learning across different views, enabling the generation of oriented bounding boxes from points. PMHO \cite{pmho} first uses SAM as a point-to-mask converter. Then, it converts the initial mask into a horizontal bounding box (HBB) and uses an HBB-to-OBB network to obtain the final oriented bounding boxes (OBB).
In our work, we aim to fine-tune the original SAM model using point annotations to better adapt it to RSIs. Consequently, a straightforward idea is to use the proposed PointSAM as a point-to-box converter, similar to PMHO. We also conducted experiments on weakly supervised oriented object detection and achieved promising results.

\begin{figure*}[tp]
	\centering
\resizebox{0.98\linewidth}{!}{\input{pic/pointsam.tex}}
	\caption{Overall architecture of the proposed PointSAM. (a) Offline prototype generation. First, feature points are obtained from the target domain dataset using the encoder of the frozen Source SAM model, and then clustering is applied to these features to obtain the target domain prototypes. (b) SAM with self-training. The training images undergo strong augmentation and weak augmentation, and are then processed through two encoders with shared weights: the teacher and the student. The original layers of the encoder are frozen, and Low-Rank Adaptation (LoRA) is used for fine-tuning. Calibration refers to Negative Prompt Calibration, which is used to obtain refined masks by adjusting the negative prompts. Matching refers to Hungarian matching, which is used to align predicted prototypes with target prototypes.
	}
	\label{fig:pointsam}
\end{figure*}

\subsection{Self-Training}
\label{sec:self-training}
Self-training is widely used in fields such as semi-supervised learning \cite{ut,st,ossod} and domain adaptation \cite{irg, ttac,sfod}. This is due to its ability to progressively assign pseudo-labels to unlabeled data, thereby enhancing the training of labeled data. This iterative process not only leverages the information present in the unlabeled data but also mitigates overfitting to the limited labeled data. However, in the absence of labeled data, self-training often falls into \textit{confirmation bias}\cite{bias}. This occurs because the model may continually reinforce its own incorrect predictions during the generation of pseudo-labels, especially when the initial pseudo-label quality is low. This bias can cause the model to gradually deviate from the correct decision boundary, ultimately affecting the overall performance of the model. There are two main approaches to address this issue: one is to use feature alignment \cite{stfar,yoo2024and}, and the other is to apply logit constraints \cite{wesam,tribe,irg} to regularize self-training. For example, STFAR \cite{stfar} uses instance-level and image-level features to align the features of the source and target domains. WeSAM \cite{wesam} uses a frozen source domain network as the anchor network to regularize the target teacher and student models.

Although these methods can mitigate error accumulation in self-training, we find the following shortcomings: 1) Feature alignment methods require the use of features from target data. Due to the large scale of SAM's pre-training data, obtaining features from the target data is unrealistic. 2) Logits-based methods often rely on the predicted logits, but if the source model cannot provide accurate predictions, these methods will not yield good results.  In our work, we directly use the features from a frozen source model on the target data as prototypes to regularize self-training. Furthermore, we only select embeddings corresponding to the labeled points from the encoder's extracted features, thus avoiding the issue of excessive logit prediction bias.

\subsection{Recognition in Remote Sensing Images}
Remote sensing images (RSIs) are captured by airborne or satellite sensors to observe and analyze the Earth's surface. These images provide critical information for a wide range of applications, including environmental monitoring \cite{9779739}, urban planning \cite{sslchange}, disaster management \cite{10551864}, and military operations \cite{10109736}. A key characteristic of RSIs is their overhead perspective, typically categorized into optical images and Synthetic Aperture Radar (SAR) images. The objects detected in these images often exhibit significant scale variations and dense distributions. Existing methods address these challenges through feature processing \cite{cad-net,psgcnet,10004975,10209224}, loss function design \cite{afdet,csl,10587005,tinet}, and post-processing stages \cite{10282477,dt-nms}. However, these approaches are primarily tailored for object counting and detection tasks, with interactive segmentation remaining relatively unexplored. In our case, SAM often struggles to segment dense objects, especially when only points are used as prompts. If the positive prompt is not well-annotated, the predicted mask may become confused with the surrounding foreground. Negative prompts can help mitigate this issue, but selecting the correct negative prompt remains challenging. Therefore, we propose using network-adaptive learned negative prompts to calibrate the predicted masks.

\section{Methodology}
\subsection{Preliminary}


\subsubsection{Segment Anything Model}
SAM \cite{sam} mainly consists of three components: an image encoder $\Phi_{\text{img}}$, a prompt encoder $\Phi_{\text{prompt}}$, and a mask decoder $\Phi_{\text{mask}}$. The image encoder is based on the Vision Transformer \cite{vit} and extracts the input image as image embeddings. The prompt encoder is used to encode various types of prompts $\mathcal{P}$, generally including points, boxes, masks, and text. There are two types of point prompts: positive prompts and negative prompts. Positive prompts are used to refer to the foreground, while negative prompts are used to refer to the background. The mask decoder is used to combine the outputs of the image encoder and the prompt encoder to generate the final mask predictions $\mathcal{M}_{pred}$. Given an input image $\mathcal{I}_{img} \in \mathbb{R}^{C \times H\times W}$, the entire process can be simplified as:
\begin{equation}
    \mathcal{M}_{pred} = \Phi_{\text{mask}}(\Phi_{\text{img}}(\mathcal{I}_{img}),\Phi_{\text{prompt}}(\mathcal{P})).
    \label{eq:sam}
\end{equation}

In the training process of SAM, ground truth masks $\mathcal{M}_{gt}$ are used for supervision.

\subsubsection{Low-Rank Adaptation}
Low-Rank Adaptation (LoRA) \cite{lora} is a technique used to reduce the computational and memory requirements of training large neural networks. By approximating weight updates with low-rank matrices, LoRA allows for more efficient fine-tuning of pre-trained models. This approach enables the adaptation of large models to new tasks or datasets with significantly lower resource consumption while maintaining performance. For each weight in the encoder network $\theta \in \mathbb{R}^{d_i \times d_o}$, we use a low-rank approximation $\omega=A B$ where $A \in \mathbb{R}^{d_i \times r}$ and $B \in \mathbb{R}^{r \times d_o}$ with $r$ indicating the rank. We can achieve a compression rate $r = \left(d_i+d_o\right) / d_i \cdot d_o$. Only $A$ and $B$ are updated via backpropagation during adaptation to reduce memory footprint. At the inference stage, the weight is reconstructed by combining the low-rank reconstruction and original weight, $\theta=\theta+A B$.



\subsection{Pointly-supervised Segment Anything Model}
In our task, only point labels are available, and there is a significant domain gap between RSIs and natural images. Therefore, our approach focuses on addressing two key challenges:  \textit{\textbf{1)} how to effectively adapt SAM to RSIs}, and \textit{\textbf{2)} how to fully exploit the potential of point annotations}. For the first challenge, we combine a self-training strategy with prototype representation constraints to ensure stable knowledge transfer and prevent the model from learning incorrect patterns. For the second challenge, we propose extracting features from point annotations to generate prototypes and refining mask predictions by adjusting negative point prompts.

The overall architecture of the Pointly-supervised Segment Anything Model (PointSAM) is illustrated in Fig.~\ref{fig:pointsam}. The pipeline is divided into two main stages: \textbf{Offline Prototype Generation} and \textbf{Self-Training with SAM}. In the first stage, we extract instance prototypes offline from the target dataset (see Sec.~\ref{sec:Offline Prototype Generation}). Target prototypes are generated through clustering and remain fixed during subsequent training. In the second stage, two augmented views of the input image $\mathcal{I}_{img}$ are generated: $\mathcal{I}_{img}^s$ with strong data augmentation and $\mathcal{I}_{img}^w$ with weak data augmentation \cite{wesam}. These augmented images are then fed into shared teacher and student encoders. While the encoder structures remain frozen, the model is fine-tuned with additional LoRA layers. Since the teacher network is dynamically trained, its image encoder extracts instance features that are stored in a memory bank \cite{moco} updated using a First-in-First-Out (FIFO) strategy to ensure that stored features remain up-to-date. These stored features are clustered to generate new predicted prototypes (see Sec.~\ref{sec:Memory Bank Updating}). The target prototypes are aligned with the predicted prototypes using a Hungarian Matching loss (see Sec.~\ref{Hungarian Matching}). Meanwhile, the teacher and student networks generate corresponding masks, $\mathcal{M}^{t}$ and $\mathcal{M}^{s}$. For $\mathcal{M}^{t}$, we apply a Negative Prompt Calibration (NPC) strategy, using positive prompts from overlapping samples as negative prompts for specified samples to refine the mask predictions. This process results in optimized masks, $\mathcal{M}^{r}$ (see Sec.~\ref{sec:Negative Prompt Calibration}), which are subsequently used as pseudo-labels to train the student network. For detailed information on the network training losses, refer to Sec.~\ref{sec:total loss}.

\subsection{Prototype-based Regularization}
General self-training methods are prone to \textit{confirmation bias} \cite{bias}. There are two common solutions to solve this problem. The first approach \cite{stfar,actmad,yoo2024and} involves aligning the predicted features extracted by the model from the source data with those extracted from the target data. However, due to the vast amount of data used to train SAM, it is challenging to obtain an accurate source distribution. Additionally, the limited number of batches used in SAM's fine-tuning can also result in inaccurate prediction distributions. Therefore, this approach is not suitable for our task. The second approach \cite{tribe,wesam} introduces an anchor model to obtain the corresponding logits to constrain the predicted logits. Specifically, this method uses the frozen weights of the source model (original SAM model) to predict the results on the target data and constrains the self-training process of the target model with these results. However, since the source model's predictions might contain significant errors, this approach may not be optimal. Instead of directly constraining predicted logits, we propose instance-level constraints without relying on source data. First, we generate target prototypes using GT points through Offline Prototype Generation. Predicted prototypes are then dynamically obtained via \textbf{Memory Bank Updating}. Finally, \textbf{Hungarian Matching} is used to align the target and predicted prototypes. 

\subsubsection{Offline Prototype Generation}
\label{sec:Offline Prototype Generation}

We begin by using the source model to extract embeddings offline for prompts corresponding to each instance in the target dataset. As illustrated in Fig. \ref{fig:pointsam}(a), given an image $\mathcal{I}_{img} \in \mathbb{R}^{C \times H \times W}$ from the target dataset, we pass it through the frozen image encoder of SAM to obtain the feature map $\mathcal{F}^{b} \in \mathbb{R}^{C_b \times H/s \times W/s}$. Given a ground truth (GT) prompt $\left(x_k, y_k\right)$ within the original image, we map it to the feature map coordinates ($x_k^{\prime} = {x_k}/s$, $y_k^{\prime} = y_k/s$) and extract the corresponding embedding $f_k^t \in \mathbb{R}^{C_b}$ from $\mathcal{F}^{b} $:
\begin{equation}
    f_k^t=\mathcal{F}^b\left(x_k^{\prime}, y_k^{\prime}\right)
\label{eq:target_feats}
\end{equation}

In this way, we can obtain sufficient feature points $\{f_k^t\}_{k=1\cdots K}$ from the source model corresponding to GT points in the target data. Next, we cluster these feature points. Since SAM is a class-agnostic segmenter, the feature points lack class labels, and the number of clusters is unknown. Consequently, directly applying KMeans for clustering is suboptimal. To overcome this limitation, we leverage the FINCH algorithm \cite{finch}, which does not require prior knowledge of the number of clusters. Subsequently, the mean feature of each cluster is computed to represent the target prototype. Let $C^t_i$ denote the $i$-th cluster:

\begin{equation} 
\mathcal{P}^{t}_i = \frac{1}{\left|C^t_i\right|} \sum_{f_k^t \in C^t_i} f_k^t 
\label{eq:target_pts}
\end{equation}

Thus, we obtain the feature prototype representations of the source model for the target dataset. Notably, $\mathcal{P}^{t}$ is not updated after extraction.

\begin{figure*}[htp]
	\centering
\resizebox{0.98\linewidth}{!}{\input{pic/neg_pt_cali.tex}}
	\caption{The process of negative prompt calibration. The positive and negative prompts are represented by red points (\textcolor{red}{\large\textbullet}) and green points (\textcolor{green}{\large\textbullet}), respectively. Different prompts input into SAM generates different initial masks. To refine these masks, an IoU matrix is calculated for each instance pair. Matrix values greater than 0 indicate that the two objects can act as negative constraints for each other. By using the positive prompt of one object as the new negative prompt for another and inputting it into SAM again, a refined mask is generated. It is worth noting that Ground Truth here refers to the mask specified by the prompt for a specific instance, not the mask for all instances.
	}
	\label{fig:neg}
\end{figure*}

\subsubsection{Memory Bank Updating}

\label{sec:Memory Bank Updating}
During SAM's self-training, we extract features associated with prompts for prototype prediction, as shown in Fig. \ref{fig:pointsam}(b). Since the teacher model provides more stable features, we use its encoder output to obtain the predicted features, following the same approach as Offline Prototype Generation. To handle the dynamic nature of network training, we use a memory bank \cite{moco} to store these features. Given a predicted instance feature $f^p_k$ generated by the teacher's image encoder with a positive prompt, we update the memory bank using the following rule, where $\mathcal{B}[0]$ is the first element in the queue and $\backslash$ indicates a removal operation.

\begin{equation}\label{eq:updating}
\vspace{-0.1cm}
    \mathcal{B}=\mathcal{B} \bigcup f_k^c,\quad 
    \mathcal{B}=\mathcal{B} \setminus \mathcal{B}[0]
\end{equation}
Here, $\mathcal{B}$ is initialized as an empty set $\mathcal{B}=\emptyset$ at the start of training. The memory bank is populated with features from the teacher model without dequeuing until $\mathcal{B}$ reaches its predefined maximum length.

The update process follows a first-in, first-out (FIFO) strategy to dynamically maintain the feature information and prevent stale features from remaining in the memory bank.

Similar to the process of generating target prototypes, we employ the FINCH \cite{finch} algorithm to cluster the features in the memory bank $\mathcal{B} = \left\{f_1^p, f_2^p, \cdots, f_K^p\right\}$. Let $C^p_j$ denote the $j$-th cluster in $\mathcal{B}$, the predicted prototypes $\mathcal{P}^p_j$ are defined as:

\begin{equation}
\mathcal{P}^{p}_j=\frac{1}{\left|C^p_j\right|} \sum_{f_k^p \in C^p_j} f_k^p
\end{equation}

\subsubsection{Hungarian Matching}
\label{Hungarian Matching}
Since the target prototypes \(\{\mathcal{P}^{t}_i\}_{i=1\cdots I}\) and predicted prototypes \(\{\mathcal{P}^{p}_j\}_{j=1\cdots J}\) cannot be directly matched one-to-one in order, a simple metric function is insufficient to enforce their consistency. Inspired by the instance matching strategy in DETR~\cite{detr}, we adopt the Hungarian Matching algorithm to compute feature similarity. 

We first define a distance matrix \(\mathbf{D} \in \mathbb{R}^{K_t \times K_p}\), where each element \(\mathbf{D}_{ij}\) represents the distance between the \(i\)-th target prototype \(\mathcal{P}^t_i\) and the \(j\)-th predicted prototype \(\mathcal{P}^p_j\). In this work, cosine similarity is used to measure the distance:
\begin{equation}
    \mathbf{D}_{ij} = 1 - \frac{\mathcal{P}^t_i \cdot \mathcal{P}^p_j}{\|\mathcal{P}^t_i\| \|\mathcal{P}^p_j\|}
\end{equation}

The Hungarian algorithm~\cite{hungarian} is then applied to solve the bipartite matching problem, finding the optimal permutation \(\pi\) that minimizes the total distance:
\begin{equation}
    \pi^* = \arg \min_{\pi \in \Pi} \sum_{i=1}^{K_t} \mathbf{D}_{i, \pi(i)}
    \label{eq:pi}
\end{equation}
where \(\Pi\) denotes the set of all possible matchings, and \(\pi(i)\) indicates the index of the predicted prototype matched to the \(i\)-th target prototype.

The final matching loss is computed as the total distance for all matched pairs:
\begin{equation}
    \mathcal{L}_{\text{match}} = \sum_{i=1}^{K_t} \mathbf{D}_{i, \pi^*(i)}
    \label{eq:match}
\end{equation}

\subsection{Negative Prompt Calibration}
\label{sec:Negative Prompt Calibration}

In SAM training, point prompts include both positive and negative prompts, which require human annotation. Positive prompts are sampled from any point within an instance, while negative prompts are more ambiguous due to the extensive background. Typically, any point outside the mask can serve as a negative prompt. However, remote sensing images present unique challenges with densely packed objects and high similarity to the background. Without boundary constraints, point supervision in self-training may lead to a single predicted mask covering multiple foreground objects or large background regions. As shown in Fig. \ref{fig:intro_pos_neg}, introducing negative prompts effectively separates objects from ambiguous regions during inference. Inspired by this, we propose a \textbf{Negative Prompt Calibration} (NPC) method that dynamically adjusts negative prompts during training.

Fig.~\ref{fig:npc_vis} illustrates the full NPC process. Given an initial set of prompt points $\mathcal{P}_{\text{init}}$, which consists of $K$ positive prompts $\mathcal{P}^{\text{pos}}_{\text{init}}$ = $\{\mathbf{p}^{\text{pos}}_k\}_{k=1 \cdots K}$ and $K$ negative prompts $\mathcal{P}^{\text{neg}}_{\text{init}}$ = $\{\mathbf{p}^{\text{neg}}_k\}_{k=1 \cdots K}$, the initial mask $\mathcal{M}^{\text{init}}_i$ for each instance is generated by feeding $\mathcal{P}_{\text{init}}$ and the encoder features into the mask encoder $\Phi_{\text{mask}}$ and the prompt encoder $\Phi_{\text{prompt}}$:
\begin{equation}
    \mathcal{M}_{\text{init}} = \Phi_{\text{mask}}(\Phi_{\text{prompt}}(\mathcal{P}_{\text{init}})).
    \label{eq:init_mask}
\end{equation}
Here, we omit the image encoder features for simplicity.

For images containing multiple objects, \( \mathcal{M}_{\text{init}} \) will also contain multiple masks. We first compute the Intersection over Union (IoU) between each pair of masks and construct an IoU matrix \(\mathbf{O}\), where each element \(\mathbf{O}_{ij}\) represents the IoU between the \(i\)-th mask \(\mathcal{M}_i\) and the \(j\)-th mask \(\mathcal{M}_j\). To exclude self-correlation, we set the diagonal elements to 0:

\begin{equation}
\mathbf{O}_{ij}= \begin{cases}
\frac{\left|\mathcal{M}_i \cap \mathcal{M}_j\right|}{\left|\mathcal{M}_i \cup \mathcal{M}_j\right|} & \text{if } i \neq j, \\
0 & \text{if } i = j.
\end{cases}
\end{equation}



For masks that intersect with a given instance mask, we identify corresponding positive prompts as candidate negative prompts. Specifically, for a given mask \(\mathcal{M}_i\), the set of candidate negative prompts \(\hat{\mathcal{P}}^{\text{neg}}\) is derived from the positive prompts of masks that intersect with \(\mathcal{M}_i\):

\begin{equation}
\hat{\mathcal{P}}^{\text{neg}} = \left\{ \mathbf{p}^{\text{pos}}_j \mid \mathbf{O}_{ij} \geq \tau_{\text{IoU}}, \; j \neq i \right\}.
\label{eq:tau_iou}
\end{equation}

We then randomly select \(k\) prompts from \(\hat{\mathcal{P}}^{\text{neg}}\) as the new negative prompts \(\tilde{\mathcal{P}}^{\text{neg}}\) for the \(i\)-th instance:

\begin{equation}
\tilde{\mathcal{P}}^{\text{neg}} \subset \hat{\mathcal{P}}^{\text{neg}}, \quad \text{with } |\tilde{\mathcal{P}}^{\text{neg}}| = k.
\end{equation}

After obtaining the new negative prompts \(\tilde{\mathcal{P}}^{\text{neg}}\), we input them along with the initial positive prompts \(\mathcal{P}^{\text{pos}}\) into SAM's mask prompt to obtain the final refined masks \(\mathcal{M}^r\):

\begin{equation}
\mathcal{M}^r = \Phi_{\text{mask}}(\Phi_{\text{prompt}}(\mathcal{P}^{\text{pos}}, \tilde{\mathcal{P}}^{\text{neg}})).
\end{equation}

In this way, the refined mask \(\mathcal{M}^r\) can be used as a pseudo-label to supervise the mask \(\mathcal{M}^s\) predicted by the student.


\subsection{Total Loss}
\label{sec:total loss}

In the original SAM model, three loss functions are used: IoU loss \(\mathcal{L}_{\text{IoU}}\), Dice loss \(\mathcal{L}_{\text{dice}}\), and Focal loss \(\mathcal{L}_{\text{focal}}\). These losses are computed between the ground truth (GT) and predicted masks. In our case, since GT masks are unavailable, these losses are used to supervise the student model's predictions \(\mathcal{M}^s\) with the refined masks \(\mathcal{M}^r\) predicted by the teacher. Additionally, we include the matching loss \(\mathcal{L}_{\text{match}}\) to enforce alignment between the target and predicted prototypes. The total loss \(\mathcal{L}_{\text{total}}\) is defined as:

\begin{equation}
    \mathcal{L}_{\text{total}} = \lambda_{\text{focal}} \mathcal{L}_{\text{focal}} + \mathcal{L}_{\text{dice}} + \lambda_{\text{match}} \mathcal{L}_{\text{match}} + \mathcal{L}_{\text{IoU}}.
    \label{equ:total_loss}
\end{equation}

\section{Experiments}
\subsection{Datasets}
To comprehensively evaluate the effectiveness of our proposed method, we conducted experiments on three widely used remote sensing instance segmentation datasets: HRSID \cite{hrsid}, NWPU VHR-10 \cite{ricnn}, and WHU \cite{whu}. The details are as follows:

\textbf{NWPU VHR-10} dataset \cite{ricnn} is a ten-class geospatial object detection dataset. It comprises 800 VHR optical remote sensing images: 715 color images sourced from Google Earth with spatial resolutions ranging from 0.5 to 2 meters, and 85 pan-sharpened color infrared images from Vaihingen data with a spatial resolution of 0.08 meters. The dataset is divided into two subsets: (a) the positive image set, containing 650 images with at least one target per image, and (b) the negative image set, consisting of 150 images with no targets. For our experiments, we selected 520 images from the positive set for training and 130 images for testing. It is worth noting that since SAM is class-agnostic, we treat all 10 categories as a single class.

\textbf{HRSID} dataset \cite{hrsid} is used for ship detection, semantic segmentation, and instance segmentation in high-resolution SAR images. It contains 5,604 high-resolution SAR ship images and 16,951 ship instances. Its spatial resolution is 0.5–3 m. It primarily consists of two scenarios: inshore and offshore. Since segmentation in the offshore scenario is relatively straightforward, we focus our experiments on the inshore dataset. Both the training and test sets exclusively use data from the inshore scenario, comprising 459 images for training and 250 images for testing. In the following text, we will refer to this as \textbf{HRSID-inshore}.
    
\textbf{WHU} dataset \cite{whu} consists of over 220,000 independent buildings extracted from aerial images with a spatial resolution of 0.075 meters and a coverage area of 450 square kilometers in Christchurch, New Zealand. We use the training set for training and the validation set for testing, with 4,736 and 1,036 images, respectively.

\subsection{Experiment Details}

\textbf{Encoder Setting:} If not otherwise specified, the image encoders used in experiments with SAM \cite{sam} and SAM2 \cite{sam2} are ViT-b and Hiera-B+, respectively.

\textbf{Prompt Generation:} For each instance mask, we randomly select \( N \) positive prompts from the corresponding GT mask and \( N \) negative prompts from outside the GT mask. We use the same method to generate prompts for both training and testing data. This practice guarantees fair evaluation of SAM which requires prompt input for segmentation. 

\textbf{Competing Methods:}
We evaluate multiple source-free domain adaptation approaches and the latest weakly supervised interactive segmentation methods. Specifically, directly testing the pre-trained model (\textbf{Direct}) with fixed prompt inputs. \textbf{TENT} \cite{tent} is a basic test-time adaptation method that adapts to the target domain by optimizing an entropy loss. \textbf{SHOT} \cite{shot} employs pseudo labels and applies a uniform distribution assumption for source-free domain adaptation. \textbf{Self-Training} \cite{st} adopt a vanilla teacher-student structure without any tricks. \textbf{Tribe} \cite{tribe} leverages anchor loss to constrain self-training. \textbf{DePT} \cite{dept} inserts visual prompts into a visual Transformer and adjusts these source-initialized prompts solely during the adaptation process without accessing the source data. \textbf{WeSAM} incorporates anchor loss and prompt-based contrastive loss into self-training.

\begin{table*}[!htb]
    \centering
    \caption{Comparison of different methods on NWPU VHR-10 test set. \textbf{Best} results are bolded, and \underline{second-best} results are underlined.}
    \small

    \begin{tabular}{lcc|cc|cc||cc|cc|cc|cc}
        \toprule
        & \multicolumn{6}{c||}{\cellcolor{lightorange}SAM-based}&\multicolumn{6}{c}{\cellcolor{lightblue}SAM2-based}\\
        \cmidrule(r){2-7} \cmidrule(r){8-13}
        Method& \multicolumn{2}{c}{\cellcolor{lightgray}1-Point} & \multicolumn{2}{c}{\cellcolor{lightgray}2-Point} & \multicolumn{2}{c||}{\cellcolor{lightgray}3-Point}& \multicolumn{2}{c}{\cellcolor{lightgray}1-Point} & \multicolumn{2}{c}{\cellcolor{lightgray}2-Point} & \multicolumn{2}{c}{\cellcolor{lightgray}3-Point}\\
        \cmidrule(r){2-3} \cmidrule(r){4-5} \cmidrule(r){6-7} \cmidrule(r){8-9} \cmidrule(r){10-11} \cmidrule(r){12-13}
        & IoU& F1& IoU& F1& IoU& F1& IoU& F1& IoU& F1& IoU& F1\\
        \midrule
        Direct test\cite{sam}& 58.06&68.80&63.93&74.92&60.98&71.95&58.28&69.43&62.68&73.87&61.76&73.39\\
        Tent\cite{tent}& 59.87&70.02&64.45&75.40&61.00&72.00&59.26&70.53&63.90&75.14&62.86&74.36\\
        Shot\cite{shot}& 61.48&72.11&65.66&\underline{76.54}&62.73&73.51&60.25&71.37&62.92&74.40&61.98&73.68\\
        Self-Training\cite{ut}& 63.94&74.11&65.34&76.05&60.47&71.94&59.62&70.38&63.63&74.36&61.86&73.27\\
        DePT\cite{dept}& \underline{64.97}&74.47&\underline{67.13}&74.35&\underline{64.92}&75.82&58.85&69.22&63.98&75.28&63.62&74.58\\
        Tribe\cite{tribe}& 64.27&73.79&64.56&75.60&60.84&71.39&\underline{61.59}&\underline{71.86}&65.54&76.05&\underline{67.02}&77.76\\
        WeSAM\cite{wesam}& 64.85&\underline{75.28}&64.86&76.00&66.03&\underline{76.73}&58.89&70.32&\underline{69.77}&\underline{79.83}&67.24&\underline{78.35}\\


        PointSAM(Ours)& \textbf{66.66}&\textbf{76.03}&\textbf{67.03}&\textbf{77.42}&\textbf{67.98}&\textbf{78.57}&\textbf{62.26}&\textbf{73.66}&\textbf{70.00}&\textbf{80.22}&\textbf{69.05}&\textbf{80.27}\\
        \midrule
        \textcolor{gray}{Supervised}& \textcolor{gray}{78.73}&\textcolor{gray}{86.74}&\textcolor{gray}{80.88}&\textcolor{gray}{88.58}&\textcolor{gray}{81.12}&\textcolor{gray}{88.79}&\textcolor{gray}{81.76}&\textcolor{gray}{88.48}&\textcolor{gray}{83.14}&\textcolor{gray}{90.11}&\textcolor{gray}{83.41}&\textcolor{gray}{90.32}\\
        \bottomrule
    \end{tabular}
    \label{tab:exp_nwpu}
\end{table*}

\textbf{Evaluation Metrics:} We report the mIoU as evaluation metrics. For each input prompt, the IoU is calculated between the ground-truth segmentation mask and the predicted mask. The mIoU averages over the IoU of all instances.

\textbf{Implementation Details}
We fine-tune the LoRA module of the image encoder using the Adam optimizer across all experiments. Training is performed on an RTX 3090 GPU with a batch size of 1, a learning rate of 0.0005, and a weight decay of 0.0001. The low-rank dimension of the LoRA module is set to 4. The coefficients \(\lambda_{\text{focal}}\) and \(\lambda_{\text{match}}\) in Eq.~\ref{equ:total_loss} are set to 20 and 0.1, respectively. For self-training, we apply both strong and weak data augmentations, following the augmentation strategies described in \cite{wesam}. Due to the presence of too many instances in some remote sensing images, to save GPU memory, we set the maximum number of training samples per image to 50.

\begin{table*}[!htb]
    \centering
    \caption{Comparison of different methods on the WHU test set. \textbf{Best} results are bolded, and \underline{second-best} results are underlined.}
    \small

    \begin{tabular}{lcc|cc|cc||cc|cc|cc}
        \toprule
        & \multicolumn{6}{c||}{\cellcolor{lightorange}SAM-based}&\multicolumn{6}{c}{\cellcolor{lightblue}SAM2-based}\\
        \cmidrule(r){2-7} \cmidrule(r){8-13}
        Method& \multicolumn{2}{c}{\cellcolor{lightgray}1-Point} & \multicolumn{2}{c}{\cellcolor{lightgray}2-Point} & \multicolumn{2}{c||}{\cellcolor{lightgray}3-Point}& \multicolumn{2}{c}{\cellcolor{lightgray}1-Point} & \multicolumn{2}{c}{\cellcolor{lightgray}2-Point} & \multicolumn{2}{c}{\cellcolor{lightgray}3-Point}\\
        \cmidrule(r){2-3} \cmidrule(r){4-5} \cmidrule(r){6-7} \cmidrule(r){8-9} \cmidrule(r){10-11} \cmidrule(r){12-13}
        & IoU& F1& IoU& F1& IoU& F1& IoU& F1& IoU& F1& IoU& F1\\
        \midrule
        Direct test\cite{sam}& 61.03&70.69&65.10&74.76&59.71&69.46&59.97&70.79&65.79&76.31&62.45&73.01\\
        Tent\cite{tent}& 61.25&70.87&65.49&75.17&59.63&69.50&60.42&71.25&65.55&76.22&62.74&73.27\\
        Shot\cite{shot}& 61.20&70.76&65.91&75.46&60.86&70.62&61.06&70.49&67.96&77.04&62.50&73.22\\
        Self-Training\cite{ut}& 64.91&73.99&68.49&77.57&59.57&69.35&65.01&75.38&68.60&78.60&68.74&77.43\\
        DePT\cite{dept}& \underline{71.31}&\underline{79.41}&73.69&81.21&\underline{73.53}&\underline{81.47}&\underline{69.52}&\underline{77.86}&\underline{74.33}&\underline{82.27}&73.91&81.88\\
        Tribe\cite{tribe}& 65.55&74.61&71.17&79.56&69.14&77.81&66.67&76.16&72.00&80.81&72.58&81.53\\
        WeSAM\cite{wesam}& 66.29&75.12&\underline{74.09}&\underline{82.07}&69.91&78.45&66.16&75.86&72.02&81.08&\underline{74.23}&\underline{82.79}\\
        PointSAM(Ours)& \textbf{72.63}&\textbf{80.39}&\textbf{76.47}&\textbf{84.10}&\textbf{77.54}&\textbf{85.23}&\textbf{73.69}&\textbf{81.21}&\textbf{76.95}&\textbf{84.55}&\textbf{75.16}&\textbf{83.91}\\
        \midrule
        \textcolor{gray}{Supervised}&\textcolor{gray}{77.15}&\textcolor{gray}{84.55}&\textcolor{gray}{79.73}&\textcolor{gray}{86.78}&\textcolor{gray}{80.54}&\textcolor{gray}{87.49}&\textcolor{gray}{78.75}&\textcolor{gray}{85.97}&\textcolor{gray}{80.40}&\textcolor{gray}{87.50}&\textcolor{gray}{88.18}&\textcolor{gray}{88.70}\\
        \bottomrule
    \end{tabular}
    \label{tab:exp_whu}
\end{table*}

        

\begin{table*}[!htb]
    \centering
    \caption{Comparison of different methods on the HRSID-inshore test set. \textbf{Best} results are bolded, and \underline{second-best} results are underlined.}
    \small

    \begin{tabular}{lcc|cc|cc||cc|cc|cc}
        \toprule
        & \multicolumn{6}{c||}{\cellcolor{lightorange}SAM-based}&\multicolumn{6}{c}{\cellcolor{lightblue}SAM2-based}\\
        \cmidrule(r){2-7} \cmidrule(r){8-13}
        Method& \multicolumn{2}{c}{\cellcolor{lightgray}1-Point} & \multicolumn{2}{c}{\cellcolor{lightgray}2-Point} & \multicolumn{2}{c||}{\cellcolor{lightgray}3-Point}& \multicolumn{2}{c}{\cellcolor{lightgray}1-Point} & \multicolumn{2}{c}{\cellcolor{lightgray}2-Point} & \multicolumn{2}{c}{\cellcolor{lightgray}3-Point}\\
        \cmidrule(r){2-3} \cmidrule(r){4-5} \cmidrule(r){6-7} \cmidrule(r){8-9} \cmidrule(r){10-11} \cmidrule(r){12-13}
        & IoU& F1& IoU& F1& IoU& F1& IoU& F1& IoU& F1& IoU& F1\\
        \midrule
        Direct test\cite{sam}& 46.56&57.46&37.80&48.34&28.32&37.57&35.40&46.14&37.26&49.07&34.89&46.75\\
        Tent\cite{tent}& 46.61&57.60&38.22&48.85&29.15&38.51&36.10&47.04&38.00&50.05&35.43&47.23\\
        Shot\cite{shot}& 47.93&58.92&40.19&50.77&28.32&37.57&35.39&46.33&37.25&48.90&33.72&45.22\\
        Self-Training\cite{ut}& 47.44&58.74&38.90&49.99&29.19&39.19&37.39&47.56&44.14&56.42&42.46&54.99\\
        DePT\cite{dept}& 50.19&61.43&\underline{43.52}&\underline{55.58}&34.73&46.08&\textbf{55.18}&\textbf{67.86}&\underline{54.76}&\underline{68.04}&\underline{54.13}&\underline{67.17}\\
        Tribe\cite{tribe}& \underline{51.22}&\underline{62.53}&42.32&53.39&32.61&42.77&42.12&55.12&46.51&59.90&39.19&51.11\\
        WeSAM\cite{wesam}& 50.50&62.43&41.95&53.58&\underline{35.51}&\underline{46.54}&47.61&60.02&47.70&60.77&45.30&59.06\\
        PointSAM(Ours)& \textbf{56.06}&\textbf{68.38}&\textbf{57.79}&\textbf{70.50}&\textbf{59.37}&\textbf{72.43}&\underline{52.45}&\underline{65.11}&\textbf{55.79}&\textbf{68.82}&\textbf{58.83}&\textbf{71.98}\\
        \midrule
        \textcolor{gray}{Supervised}& \textcolor{gray}{63.29}&\textcolor{gray}{75.32}&\textcolor{gray}{65.89}&\textcolor{gray}{77.65}&\textcolor{gray}{66.70}&\textcolor{gray}{78.50}&\textcolor{gray}{67.45}&\textcolor{gray}{78.56}&\textcolor{gray}{70.83}&\textcolor{gray}{81.61}&\textcolor{gray}{71.72}&\textcolor{gray}{82.42}\\
        \bottomrule
    \end{tabular}
    \label{tab:exp_hrsid}
\end{table*}

\definecolor{deepgreen}{rgb}{0.0, 0.5, 0.0}

\begin{table*}[!htb]
    \centering
    \caption{Ablation studies of the proposed PointSAM on the HRSID-inshore dataset. ST, PBR, and NPC refer to self-training, prototype-based regularization, and negative prompt calibration, respectively.}
    \footnotesize
    \begin{tabular}{ccc|cc|cc|cc}
        \toprule
        \multirow{2.5}{*}{\centering ST} & \multirow{2.5}{*}{\centering PBR} & \multirow{2.5}{*}{\centering NPC} & \multicolumn{2}{c}{\cellcolor{lightgray}1-Point} & \multicolumn{2}{c}{\cellcolor{lightgray}2-Point} & \multicolumn{2}{c}{\cellcolor{lightgray}3-Point}\\
        \cmidrule(r){4-5} \cmidrule(r){6-7} \cmidrule(r){8-9}
        &&& IoU & F1 & IoU & F1 & IoU & F1 \\
        \midrule
        &&& 46.56 & 57.46 & 37.80 & 48.34 & 28.32 & 37.57 \\
        \CheckmarkBold &&& 47.44 {\scriptsize\textcolor{deepgreen}{(+0.88)}} & 58.74 {\scriptsize\textcolor{deepgreen}{(+1.28)}} & 38.90 {\scriptsize\textcolor{deepgreen}{(+1.10)}} & 49.99 {\scriptsize\textcolor{deepgreen}{(+1.65)}} & 29.19 {\scriptsize\textcolor{deepgreen}{(+0.87)}} & 39.19 {\scriptsize\textcolor{deepgreen}{(+1.62)}} \\
        \CheckmarkBold & \CheckmarkBold && 53.86 {\scriptsize\textcolor{deepgreen}{(+6.30)}} & 66.40 {\scriptsize\textcolor{deepgreen}{(+8.94)}} & 50.30 {\scriptsize\textcolor{deepgreen}{(+12.50)}} & 62.42 {\scriptsize\textcolor{deepgreen}{(+14.08)}} & 48.04 {\scriptsize\textcolor{deepgreen}{(+18.85)}} & 61.20 {\scriptsize\textcolor{deepgreen}{(+23.63)}} \\
        \CheckmarkBold && \CheckmarkBold & 52.86 {\scriptsize\textcolor{deepgreen}{(+6.30)}} & 65.29 {\scriptsize\textcolor{deepgreen}{(+7.83)}} & 54.55 {\scriptsize\textcolor{deepgreen}{(+16.75)}} & 67.06 {\scriptsize\textcolor{deepgreen}{(+18.72)}} & 53.34 {\scriptsize\textcolor{deepgreen}{(+25.02)}} & 66.77 {\scriptsize\textcolor{deepgreen}{(+29.20)}} \\
        \CheckmarkBold & \CheckmarkBold & \CheckmarkBold & 56.06 {\scriptsize\textcolor{deepgreen}{(+9.50)}} & 68.38 {\scriptsize\textcolor{deepgreen}{(+10.92)}} & 57.79 {\scriptsize\textcolor{deepgreen}{(+19.99)}} & 70.50 {\scriptsize\textcolor{deepgreen}{(+22.16)}} & 59.37 {\scriptsize\textcolor{deepgreen}{(+31.05)}} & 72.43 {\scriptsize\textcolor{deepgreen}{(+34.86)}} \\
        \bottomrule
    \end{tabular}
    \label{tab:ablation}
\end{table*}

\subsection{Quantitative Evaluations}
We conducted quantitative evaluations on three datasets: NWPU VHR-10, WHU, and HRSID-inshore. All comparison methods were reproduced using both SAM \cite{sam} and SAM2 \cite{sam2}. We compared the IoU and F1 scores across different numbers of points, ranging from 1 to 3.

\subsubsection{NWPU VHR-10}
We first present the results of adapting various methods to the NWPU VHR-10 test set, as shown in Tab.\ref{tab:exp_nwpu}. Due to the substantial differences in viewing angles between aerial images and natural images, a significant distribution shift occurs, posing challenges for model generalization. As a result, we observe a notable performance gap between the \textbf{Supervised} upper bound and the \textbf{Direct test} baseline, with IoU differences consistently around 20\% across various numbers of prompts. In contrast, \textbf{Ours} consistently achieves the highest performance across both IoU and F1 metrics compared to other methods. Although \textbf{Tent} and \textbf{Shot} methods have shown promising results in image-level tasks, segmentation tasks operate at the pixel level, which introduces greater complexity. Self-training-based methods (\textbf{Tribe}, \textbf{DePT}, and \textbf{WeSAM}) each exhibit distinct strengths, and all outperform the original self-training methods. This highlights the crucial role of regularization in network training, especially under weak supervision conditions.  We also find that \textbf{SAM2} outperforms \textbf{SAM} in both direct test and supervised settings, demonstrating its superior generalization capability. However, when SAM2 is integrated into other methods, the performance improvement over SAM varies. This inconsistency arises because, despite incorporating SAM2, we continued to use SAM's approach in integrating. The unique advantages of SAM2's memory module were not fully utilized, which presents an opportunity for further exploration in future work.

\subsubsection{WHU}
Building extraction is highly practical in remote sensing image processing. The irregular shapes of buildings as captured from overhead views introduce significant challenges for direct testing with SAM. As shown in Tab. \ref{tab:exp_whu}, \textbf{Direct test} with SAM or SAM2 shows a performance gap exceeding 10\% compared to the \textbf{supervised} method. Our approach effectively narrows this gap to within 5\%. This is because, although the shapes of buildings vary, their contours are distinct. \textbf{Ours} effectively adapts the source domain to the target domain. It can be observed that the performance of the \textbf{Self-Training} method decreases as the number of points increases. This is because semantically ambiguous points lead to cumulative errors in the training. \textbf{DePT} and \textbf{WeSAM} show significant improvements compared to self-training; however, they are not consistently effective in all cases.

\subsubsection{HRSID-inshore}
Unlike optical images, SAR images present a larger domain gap. Additionally, imaging conditions can lead to ships appearing hollow or introducing significant noise. As shown in Tab. \ref{tab:exp_hrsid}. It can be observed that the \textbf{Direct test} performance differs significantly from the \textbf{Supervised} performance, with a gap of up to 40\% in the 3-point setting. Additionally, increasing the number of prompts does not necessarily enhance performance. As the number of points increases, suboptimal positive prompts may have a greater negative impact on performance. For example, most methods that use SAM as the base model experience a decline in performance as the number of prompts increases. Even with the more advanced SAM2, this limitation cannot be fully addressed. In contrast, \textbf{Ours} consistently improves both IoU and F1 scores under the same conditions except for being slightly lower than \textbf{DePT} in the 1-point setting. This is because the proposed NPC strategy adjusts the negative prompts to appropriate positions, allowing the positive prompts to generate more accurate masks.

\subsection{Ablation study}

\subsubsection{Impact of different components}

In this section, we analyze the effectiveness of individual components on the HRSID-inshore dataset. As shown in Table \ref{tab:ablation}, the first row represents the \textit{baseline}, where the vanilla SAM \cite{sam} is tested directly. When \textit{Self-Training} (ST) is introduced, there is only a slight improvement, as the strong and weak data augmentations enhance the network's robustness but cannot prevent error accumulation. Adding \textit{Prototype-Based Regularization} (PBR) to self-training results in significant improvements across all metrics, with increases ranging from 10\% to 20\%. This is because regularization helps alleviate error accumulation in the network. However, when more points are used, the results still decline. This is due to the small size of the targets, where additional points may appear on object boundaries, leading to misclassification of background as foreground. Adding \textit{Negative Prompt Calibration} (NPC) to self-training effectively addresses this issue. It maintains stable results for each point setting and significantly improves performance over ST. When both NPC and PBR are incorporated, the performance reaches its best across all metrics. Especially in the 3-point setting, performance shows more than a 30\% improvement compared to the baseline. This suggests that the two strategies are not mutually exclusive and can complement each other.






\subsubsection{Alternative Distance Metric in Hungarian Match}
The Tab. \ref{tab:exp_metric} shows the performance of different distance metrics (Cosine, L1, L2) in Eq.\ref{eq:pi} under 1-, 2-, and 3-point settings. The cosine metric performs best in all cases. This is due to it focus on direction similarity rather than absolute magnitude, as well as its advantages in handling sparse and high-dimensional data.

\begin{table}[!htb]
    \centering
    \caption{The impact of different distance metrics in Hungarian matching on the HRSID-inshore dataset.}
    \footnotesize
    \definecolor{MyRed}{RGB}{255,0,0}
    \definecolor{MyBlue}{RGB}{0,0,255}
    \begin{tabular}{c|cc|cc|cc}  
        \toprule
        \multirow{2.5}{*}{Distance Metric} & \multicolumn{2}{c}{1-Point} & \multicolumn{2}{c}{2-Point} & \multicolumn{2}{c}{3-Point} \\
        \cmidrule(r){2-3} \cmidrule(r){4-5} \cmidrule(r){6-7} 
        & IoU & F1 & IoU & F1 & IoU & F1 \\
        \midrule
        Cosine& \textbf{56.06} & \textbf{68.38} & \textbf{57.79} & \textbf{70.50} & \textbf{59.37} & \textbf{72.43} \\
        L1& 54.87 & 67.81 &56.57 &69.36 & 58.14 &71.10 \\
        L2& 55.42 & 68.04 &56.14 &68.90& 58.43 & 71.43 \\

        \bottomrule
    \end{tabular}

    \label{tab:exp_metric}
\end{table}

\begin{figure}[htbp]
    \centering
    \begin{tabular}{cc}
        \includegraphics[width=0.8\linewidth]{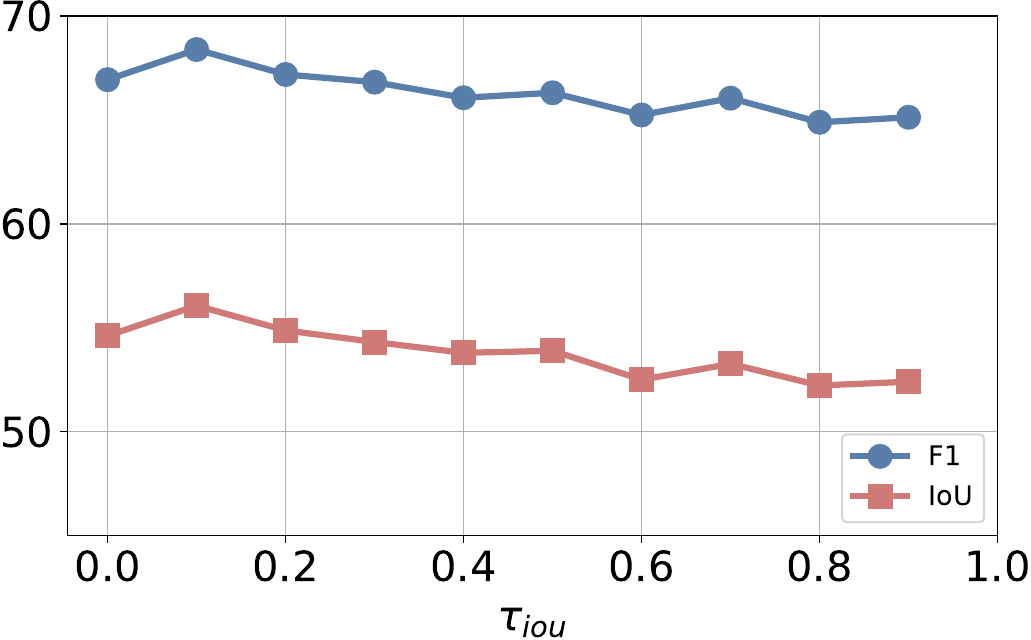} 
    \end{tabular}
    \caption{
The impact of different thresholds of IoU on the HRSID-inshore dataset with 1-point.}
    \label{fig:exp_iou_thr}
\end{figure}

\subsubsection{Alternative IoU threshold}
As mentioned earlier, NPC utilizes the IoU between masks to determine whether to use them as negative prompts. Hence, we evaluated the impact of different IoU thresholds in Eq.\ref{eq:tau_iou} on the HRSID-inshore dataset, selecting values from 0 to 0.9 at intervals of 0.1. As shown in Fig. \ref{fig:exp_iou_thr}, the results peak at a threshold of 0.1. When the threshold is set to 0, the performance is slightly lower, likely due to the introduction of noisy prompts. As the IoU threshold increases beyond 0.1, both F1 and IoU metrics exhibit a downward trend. This decline is attributed to the reduced likelihood of negative prompt adjustments at higher thresholds, diminishing the influence of NPC.

\begin{figure}[htbp]
    \centering
    \begin{tabular}{cc}
        \includegraphics[width=0.8\linewidth]{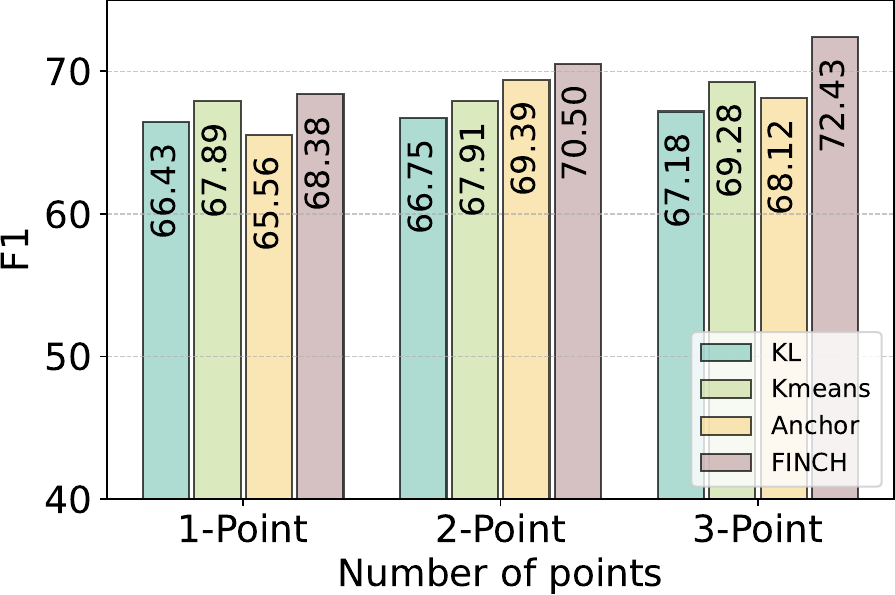} 
    \end{tabular}
    \caption{
The comparison of the different feature clustering and alignment methods on the HRSID-inshore dataset.}
    \label{fig:align_comparison}
\end{figure}
\subsubsection{Comparison with other feature alignment methods.}
We compared different clustering and feature alignment methods, and the results are shown in Fig. \ref{fig:align_comparison}. \textbf{KLD} constrains the feature mean and variance of the source and target models on the target data using the Kullback-Leibler divergence. \textbf{Kmeans} refers to using the Kmeans algorithm for feature clustering in PBR. \textbf{Anchor} denotes keeping NPC unchanged and replacing PBR with the anchor model from WeSAM\cite{wesam}. \textbf{FINCH} is the clustering method adopted in this study. The results demonstrate that \textbf{FINCH} outperforms other methods across various point settings. Due to insufficient data, \textbf{KLD} leads to inaccurate variance estimation and performs poorly. \textbf{Kmeans} performs slightly worse than \textbf{FINCH} because it requires manually setting fixed clustering centers, which are not adaptive to the feature distribution. Moreover, its computational speed is over three times slower than \textbf{FINCH}. The performance of the anchor model is inconsistent across the three different prompt quantities, as it is susceptible to inaccurate logit predictions.

\begin{figure}[htbp]
    \centering
    \begin{tabular}{cc}
        \includegraphics[width=0.8\linewidth]{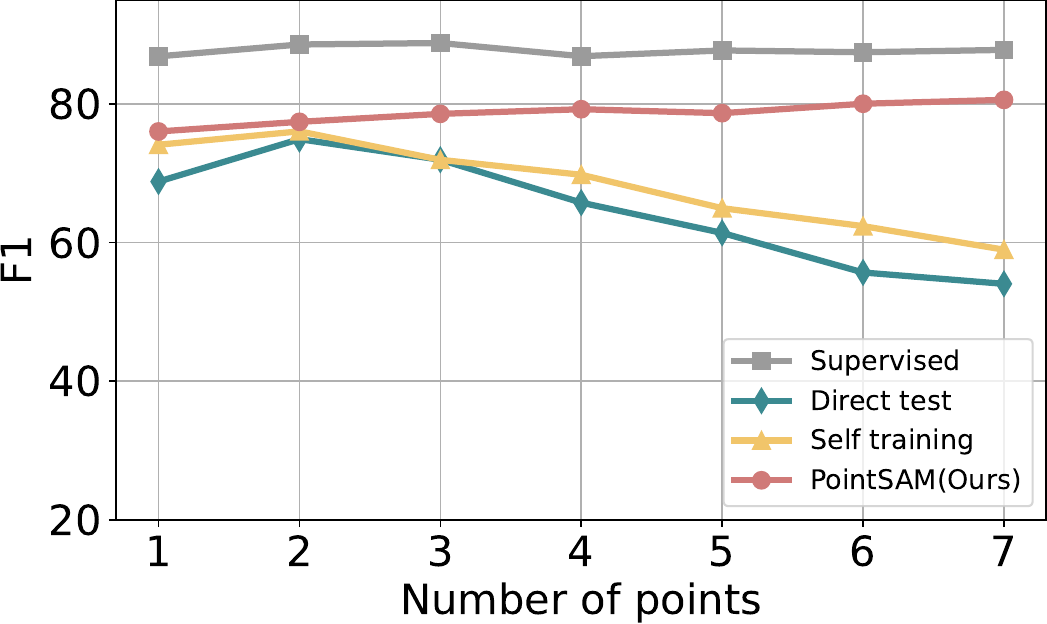} 
    \end{tabular}
    \caption{
The impact of the number of points on different methods on the NWPU VHR-10 dataset.}
    \label{fig:exp_points}
\end{figure}

\begin{figure*}[!htb]
    \centering
    \begin{tabular}{@{}c}
        \includegraphics[width=0.98\linewidth]{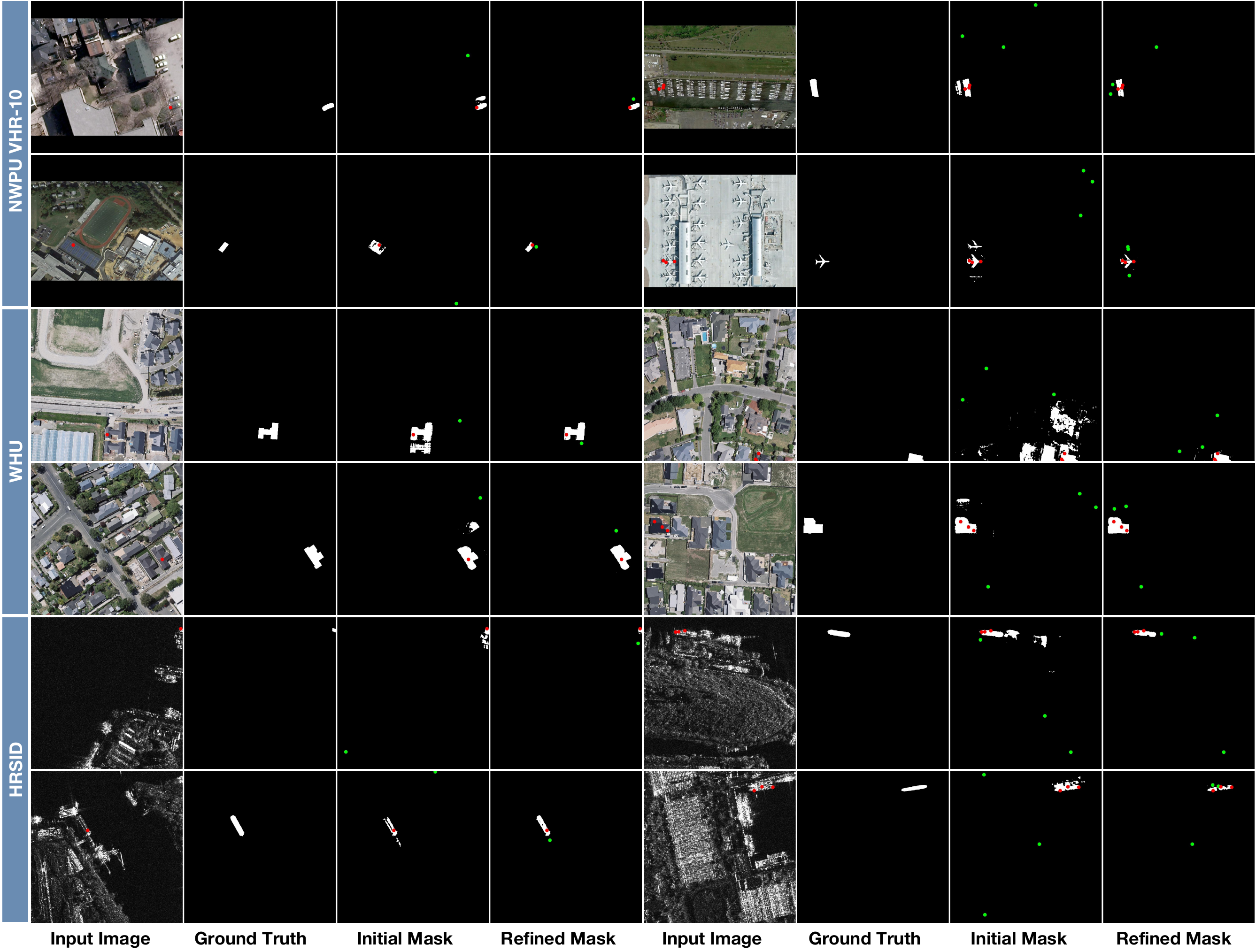}\\
        
    \end{tabular}
    \caption{
Visualization of negative prompt calibration results during training. Positive prompts are marked with red points (\textcolor{red}{\large\textbullet}), while negative prompts are marked with green points (\textcolor{green}{\large\textbullet}). The refined mask is obtained by applying negative prompt calibration to the initial mask.  This calibration effectively guides the negative prompts to more precise regions, resulting in improved mask predictions.}
    \label{fig:npc_vis}
\end{figure*}

\subsubsection{Alternative fine-tuning methods}
We compared the performance impact of different fine-tuning methods, such as \textbf{LoRA}, \textbf{Adapter}, and \textbf{LayerNorm}. \textbf{LoRA} enables efficient fine-tuning by introducing low-rank matrices into the original model, adding only minimal parameters \cite{lora}. The \textbf{Adapter} method inserts lightweight modules into the model layers, allowing task-specific learning without modifying the original parameters; in our experiments, we selected the adapter from this work \cite{wu2023medical}. \textbf{LayerNorm} fine-tunes only the LayerNorm parameters of the original SAM \cite{lei2016layer}. As shown in Table \ref{tab:finetune_method}, all fine-tuning methods perform better than direct testing, except for the \textbf{Adapter} in the 1-point case. \textbf{LoRA} outperforms all other fine-tuning methods across all metrics. The \textbf{Adapter} and \textbf{LayerNorm} methods are relatively limited as they are overly focused on local adjustments, which prevents them from achieving global adaptability in complex remote sensing images.

\begin{table}[!htbb]
    \centering
    \caption{The impact of different fine-tuning methods on the HRSID-inshore dataset.}
    \footnotesize
    \begin{tabular}{l|cc|cc|cc}  
        \toprule
        \multirow{2.5}{*}{Fine-tuning Method} & \multicolumn{2}{c}{1-Point} & \multicolumn{2}{c}{2-Point} & \multicolumn{2}{c}{3-Point} \\
        \cmidrule(r){2-3} \cmidrule(r){4-5} \cmidrule(r){6-7} 
        & IoU & F1 & IoU & F1 & IoU & F1 \\
        \midrule
        Direct Test & 46.56 & 57.46 & 37.80 & 48.34 & 28.32 & 37.57 \\
        \midrule
        LoRA \cite{lora} & \textbf{56.06} & \textbf{68.38} & \textbf{57.79} & \textbf{70.50} & \textbf{59.37} & \textbf{72.43} \\
        Adapter \cite{wu2023medical} & 40.06 & 53.15 & 47.83 & 61.30 & 47.09 & 61.27 \\
        LayerNorm \cite{lei2016layer} & 49.26 & 61.39 & 53.19 & 66.33 & 38.01 & 50.71 \\
        \bottomrule
    \end{tabular}
    \label{tab:finetune_method}
\end{table}

\subsubsection{How about more points?}
We validated the results of different methods under an increased number of point prompts. As shown in Fig. \ref{fig:exp_points}, simply adding more points does not consistently lead to better performance. This is because increasing the number of points also raises the likelihood of including low-quality points. Such noise can negatively affect the segmentation results of other points. For the \textbf{Supervised} method, the results remained relatively unchanged due to the presence of full-mask constraints. \textbf{Direct test} achieved its best results with two points; however, as the number of points increased, the F1 score gradually decreased. Similarly, \textbf{Self-training} showed a decline in results due to the generation of noisy pseudo-labels. In contrast, our proposed \textbf{PointSAM} maintained stable results, approaching the performance of \textbf{Supervised} one. This is because negative prompt calibration effectively corrected the prompts and reduced the impact of inaccurate masks caused by too many points.

\begin{figure*}[htbp]
    \centering
    \begin{tabular}{@{}c}
        \includegraphics[width=0.98\linewidth]{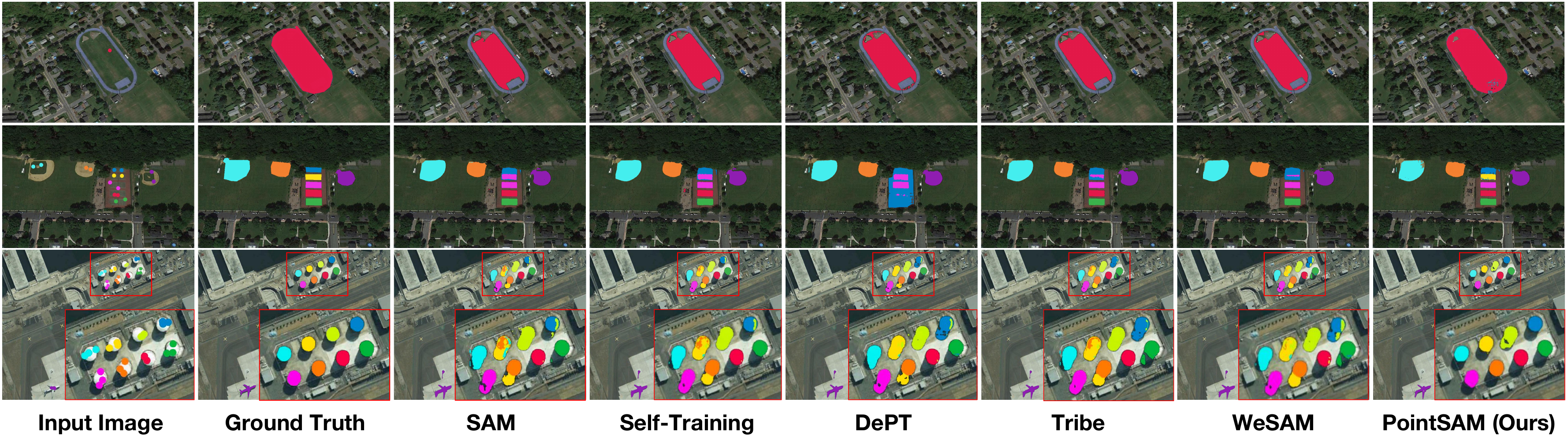}\\
        
    \end{tabular}
    \caption{Comparative results of different methods on the NWPU VHR-10 dataset. Rows 1-3 present the results under 1, 2, and 3 prompts, respectively. Zoomed-in regions in the images are used to highlight the detail.}
    \label{fig:vis_nwpu}
\end{figure*}

\begin{figure*}[htbp]
    \centering
    \begin{tabular}{@{}c}
        \includegraphics[width=0.98\linewidth]{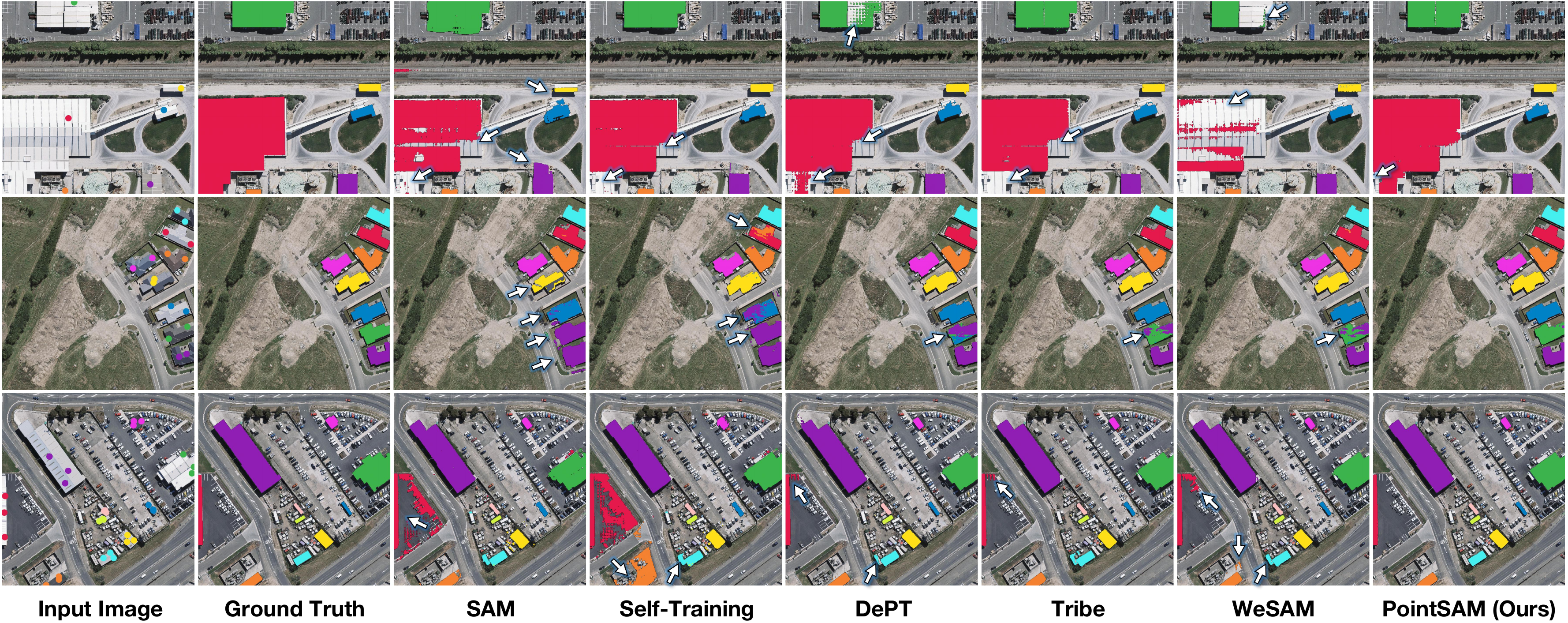}\\
        
    \end{tabular}
    \caption{Comparative results of different methods on the WHU dataset. Rows 1-3 present the results under 1, 2, and 3 prompts, respectively. White arrows in the images are used to highlight the detail.}
    \label{fig:vis_whu}
\end{figure*}

\begin{figure*}[htbp]
    \centering
    \begin{tabular}{@{}c}
        \includegraphics[width=0.98\linewidth]{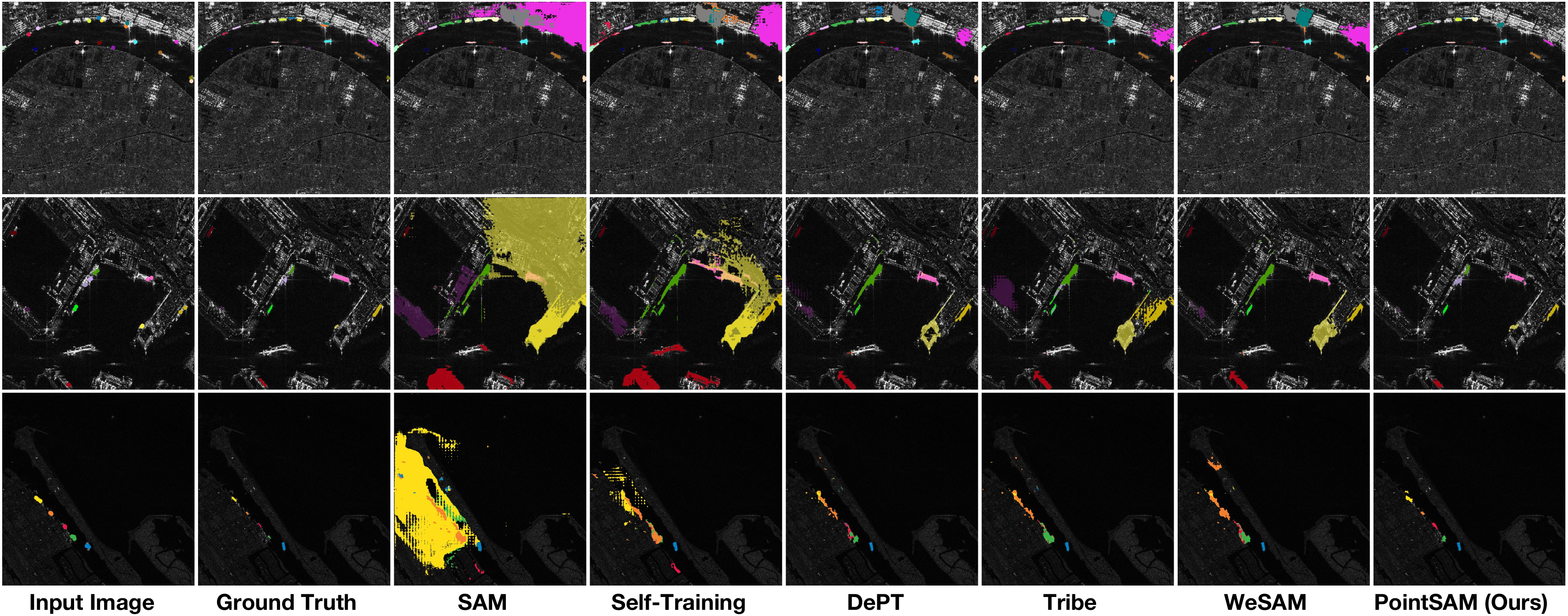}\\
        
    \end{tabular}
    \caption{Comparative results of different methods on the HRSID-inshore dataset. Rows 1-3 present the results under 1, 2, and 3 prompts, respectively.}
    \label{fig:vis_hrsid}
\end{figure*}

\subsection{Qualitative Evaluations}
\subsubsection{Visualization of the NPC Process}
To visually demonstrate the impact of NPC during training, we present the progression from initial masks to refined masks across three datasets, as shown in Fig. \ref{fig:npc_vis}. Red and green points denote positive and negative prompts, respectively. In the NWPU VHR-10 dataset, the subtle texture of tennis courts causes initial masks to cover adjacent courts. Positive prompts near object edges, such as cars, are often affected by nearby objects. By treating overlapping targets as negative prompts, redundant masks are effectively removed. As the number of positive prompts increases, prompts near object boundaries, such as airplanes or harbors, tend to cause semantic ambiguity. This ambiguity is resolved by NPC. In the WHU dataset, buildings with similar colors often result in interference in masks generated from prompts. Given the high density of buildings in each image, the method identifies ambiguous masks nearby and refines the boundaries of the target mask. The HRSID inshore dataset is the most challenging due to the SAR imaging mechanism. Ships and the surrounding scenes share nearly identical colors. Additionally, the targets are small and may have hollow structures. If constraints are applied to each negative prompt, non-target regions are likely to be included in the mask. Despite these challenges, the method suppresses redundant regions effectively, regardless of the number of prompts.

\subsubsection{Visualization of results from different methods}
We then present the comparative results of different methods across various datasets in Fig. \ref{fig:vis_nwpu}, Fig. \ref{fig:vis_hrsid}, and \ref{fig:vis_whu}. Rows 1-3 show the results with 1 to 3 prompts. It can be observed that due to the bird's-eye view in remote sensing images, there is a significant difference from natural images. Directly using the original SAM leads to an inability to distinguish each target clearly. For example, in the third row in Fig.\ref{fig:vis_whu}, the white building on the left and the parking lot on the right are treated as the same object. Even more notably, in the HRSID-inshore dataset (see Fig.\ref{fig:vis_hrsid}), most of the inshore areas are misidentified as positive masks. Self-training transfers the source model to the target data, reducing more redundant areas and producing relatively more complete predicted masks compared to direct testing. However, it still fails to mitigate the interference between adjacent objects, such as the tennis court in the second row in Fig. \ref{fig:vis_nwpu} and the building in the fifth row in Fig. \ref{fig:vis_whu}. DePT, Tribe, and WeSAM are all improvements based on self-training, and they handle mask details better than self-training. However, in more challenging scenarios, they still fail to achieve optimal results.  For instance, in the third row of Fig. \ref{fig:vis_nwpu}, the storage tank and its shadow remain indistinguishable, and ships in inshore scenes are not accurately segmented (Fig. \ref{fig:vis_hrsid}). In contrast, our method excels at handling objects in dense scenes, achieving performance close to the ground truth.

\subsection{PointSAM as a Detection Box Generator} 
In this section, we serve PointSAM as a point-to-box generator. PointSAM can generate corresponding masks based on points, and by calculating the minimum enclosing rectangle of the mask, we can obtain the corresponding horizontal bounding box (HBB). These HBBs can then be fed into a detector that converts horizontal boxes to rotated boxes, achieving \textit{point-supervised oriented object detection}. To validate the effectiveness of this approach, we conducted experiments on the HRSID dataset, which includes both inshore and offshore scenarios. All experiments were conducted with an input size of 800×800, running for 12 epochs, and using ResNet-50 as the backbone. As shown in Table 6, we compared our method with representative algorithms based on OBB supervision, HBB supervision, and point supervision. It can be observed that the H2RBox-v2 and the method proposed by \citet{10463064} based on HBB can achieve performance comparable to OBB supervision. The poor performance of H2RBox may be attributed to the large number of small objects in the HRSID dataset. Therefore, our approach also utilizes H2RBox-v2 as the detector for converting HBB to OBB. Compared to vanilla SAM, our method achieves a 15\% improvement.
This is because directly using SAM can result in unclear segmentation masks for objects in dense scenes, which in turn leads to inaccuracies in the minimum enclosing rectangles. Similarly, our method slightly outperforms Point2Rbox. Essentially, both Ours and Point2RBox leverage prior knowledge to learn the size information of the targets. There remains a gap of nearly 20\% compared to the HBB-supervised methods. Future work could focus on integrating multiple types of priors to bridge this gap.

\begin{table}
    \centering
    \caption{Comparisons results of different detectors based on HRSID.}
    \footnotesize  
    \begin{tabular}{l|c|c|c}
        \toprule
        Methods  & Backbone  & Recall(\%) & AP$_{50}$(\%)  \\
        \midrule
        \multicolumn{4}{l}{\textbf{\emph{OBB-supervised}}}    \\
        \midrule
        FCOS-O$^*$ \cite{fcos}   & ResNet-50  & 83.4  & 78.4 \\
        Faster RCNN-O$^*$ \cite{frcn}   & ResNet-50 & 83.1 & 78.0 \\
        RetinaNet-O \cite{kld}  & ResNet-50  & 80.2  & 72.3    \\
        Oriented R-CNN \cite{xie2024oriented}    & ResNet-50 & 85.0    & 79.9 \\
        \midrule
        \multicolumn{4}{l}{\textbf{\emph{HBB-supervised}}}    \\
        \midrule
        H2RBox \cite{h2rbox}    & ResNet-50 & 47.6    & 24.3\\
        H2RBox-v2 \cite{h2rboxv2}    & ResNet-50 &81.6   & 76.5\\
        \citet{10463064} & ResNet-50 & 85.0    & 81.5 \\
        \midrule
        \multicolumn{4}{l}{\textbf{\emph{Pointly-supervised}}}    \\
        \midrule
        Point2RBox \cite{point2rbox}  & ResNet-50 & 64.2    & 57.1 \\   
        SAM + H2RBox-v2  & ResNet-50 & 56.6   & 44.7 \\   
        \rowcolor{lightgray} 
        PointSAM + H2RBox-v2 (Ours)    & ResNet-50 &68.9  & 59.5\\
        \bottomrule
    \end{tabular}
\end{table}

\section{Conclusion}
In this paper, we propose PointSAM, which adapts vanilla SAM to remote-sensing images using only point labels. Our method is based on a self-training framework. The proposed prototype-based regularization overcomes the issue of error accumulation in self-training by aligning prototypes predicted by the source and target models using the Hungarian matching algorithm. Negative prompt calibration effectively addresses the problem of densely distributed objects in RSIs by leveraging the spatial adjacency relationships of instances. Our method outperforms comparison algorithms on three widely used RSI datasets, NWPU VHR-10, HRSID, and WHU, and approaches the performance of supervised methods. Additionally, we also utilize the proposed PointSAM as a point-to-box generator to train a rotated box detector, achieving promising results. However, our method still has some issues to be improved. On the one hand, the self-training-based approach uses a dual-branch structure, which can result in slower training speeds. On the other hand, negative prompt calibration does not work well for objects with sparse distributions. Therefore, further consideration could be given to integrating information between images to effectively distinguish between foreground and background.

\small
\bibliography{refs/ref}

\end{document}